\documentclass{article}
\pdfoutput=1

\usepackage[preprint,nonatbib]{neurips_2021}

\usepackage[utf8]{inputenc}
\usepackage[T1]{fontenc}    %
\usepackage{hyperref}       %
\hypersetup{
    colorlinks=true,
    citecolor=blue,
    linkcolor=blue,
    urlcolor=magenta,
}
\usepackage{url}            %
\usepackage{amsfonts}       %
\usepackage{nicefrac}       %
\usepackage{microtype}      %
\usepackage{xcolor}         %

\usepackage{booktabs}
\usepackage{subcaption}
\usepackage{graphicx}

\usepackage{biblatex}
\bibliography{refs}

\definecolor{train}{RGB}{27,158,119}
\definecolor{valid}{RGB}{217,95,2}
\definecolor{test}{RGB}{117,112,179}
\definecolor{cnndm}{RGB}{231,41,138}
\newcommand{\trainkey}{\textbf{\color{train}Train}}
\newcommand{\validkey}{\textbf{\color{valid}Valid}}
\newcommand{\testkey}{\textbf{\color{test}Test}}
\newcommand{\cnndmkey}{\textbf{\color{cnndm}CNN/DM}}
\newcommand{\idealkey}{\textbf{Ideal}}

\newcommand{\validlongform}{TL;DR Validation}
\newcommand{\redditlongform}{TL;DR Test}
\newcommand{\cnndmlongform}{CNN/DM Test}

\newcommand{\calibrationlegendcommon}{over all points in the \trainkey{}ing, \validkey{}ation and \testkey{} TL;DR datasets and the out-of-distribution \cnndmkey{} dataset. We then group predictions into $10$ bins based on the model's confidence ($x$-axis), i.e.\ the probability of the most likely class, and compute the mean accuracy ($y$-axis) within that bin relative to human labels. Lines closer to the \idealkey{} black dashed line $y = x$ indicate a better calibrated model, with points above or below $y = x$ indicating the model is respectively under-confident or over-confident.}

\definecolor{random}{RGB}{27,158,119}
\definecolor{class}{RGB}{217,95,2}
\definecolor{thompson}{RGB}{117,112,179}
\definecolor{variance}{RGB}{231,41,138}
\newcommand{\randomkey}{\textbf{\color{random}Random}}
\newcommand{\uncertaintykey}{\textbf{\color{class}Uncertainty}}
\newcommand{\thompsonkey}{\textbf{\color{thompson}Thompson}}
\newcommand{\variancekey}{\textbf{\color{variance}Variance}}

\newcommand{\activelegendcommon}{Shaded regions are 95\% confidence intervals, computed via bootstrap over results from 5 random training seeds. \textbf{Key}: \randomkey{} samples points randomly; \uncertaintykey{} chooses the point which the model is least confident about; \thompsonkey{} picks an ensemble member uniformly at random and selects the point which that member predicts the highest reward for; \variancekey{} selects points with maximal variance between ensemble members.}

\definecolor{nobootstrap}{RGB}{27,158,119}
\definecolor{bootstrap}{RGB}{217,95,2}

\newcommand{\nobootstrapkey}{\textbf{\color{nobootstrap}without}}
\newcommand{\bootstrapkey}{\textbf{\color{bootstrap}with}}

\newcommand{\uncertaintylegendhead}{We report the Spearman correlation between model error and estimated epistemic uncertainty (y-axis), against the number of ensemble members ($x$-axis) \bootstrapkey{} and \nobootstrapkey{} bootstrapping} 
\newcommand{\uncertaintylegendtail}{We measure model error by the KL divergence between the model's predicted distribution, and the oracle distribution. We estimate epistemic uncertainty by the variance between ensemble members.}

\title{Uncertainty Estimation for Language Reward Models}
\author{
    Adam Gleave\textsuperscript{1}\thanks{Work conducted during an internship at DeepMind.}\quad%
    Geoffrey Irving\textsuperscript{2}
}
\date{}

\begin{document}

\maketitle
\begin{center}
\vspace{-2.8em}
\textsuperscript{1}UC Berkeley \quad \textsuperscript{2}DeepMind \\
\texttt{gleave@berkeley.edu}
\end{center}

\begin{abstract}

Language models can learn a range of capabilities from unsupervised training on text corpora.
However, to solve a particular problem (such as text summarization) it is typically necessary to \emph{fine-tune} them on a task-specific dataset.
It is often easier for humans to choose between options than to provide labeled data, and prior work has achieved state-of-the-art performance by training a reward model from such preference comparisons.
However, collecting a large preference comparison dataset is still expensive---and the learned reward models are unreliable out-of-distribution.
We seek to address these problems via uncertainty estimation, which can improve sample efficiency and robustness using active learning and risk-averse reinforcement learning (RL).
Specifically, we use bootstrap aggregating (bagging) to train an ensemble of reward models differing in the initialization of their final layer.
Ensembles have proved successful in prior applications of active learning~\parencite{christiano:2017,beluch:2018}, but we find that in our setting ensemble active learning does not outperform random sampling.
Further experiments show that while the aggregate predictions are well-calibrated, the ensemble's estimated epistemic uncertainty is only weakly correlated with model error.
We suspect this is because the ensemble members are fine-tuned from a single model and so are similar to one another.
This suggests current pre-training methods will need to be modified to support uncertainty estimation, e.g.\ by training multiple language models.
\end{abstract}

\section{Introduction}
The objective for many natural language tasks is subtle, and difficult or impossible to procedurally specify.
For example, what makes a summary of an article ``good''?
It should be truthful, but must also omit some information to stay succinct.
This requires a value judgement: what part of the article is \emph{most important} to the reader?
It is even more challenging to specify the objective for open-ended tasks such as conversational agents interacting with users.

A common approach is to use supervised learning to mimic some reference text written by a human~\parencite{see:2017,paulus:2018,khandelwal:2019}.
However, writing reference examples is slow and expensive.
It is possible to reuse existing data sources, but they are often low quality or an imperfect match for the task.

An alternative approach is to learn a \emph{reward model} from human feedback that evaluates the quality of textual output.
One can then use reinforcement learning (RL) to fine-tune a language model to generate output that scores highly under this reward model.
This method has been applied to perform stylistic continuation~\parencite{ziegler:2019} and summarize documents~\parencite{stiennon:2020}.

There are usually multiple settings of the model parameters that make similar predictions on the (finite) training dataset, but may disagree on novel inputs.
Prior work has ignored this and learned a single point-estimate for the parameters.
But capturing this model uncertainty could improve sample efficiency via active learning, increase robustness of the RL training procedure, and aid in quality control.
In this paper, we explore some initial approaches to estimate model uncertainty.

A key application of uncertainty estimation is active learning: choosing what textual output to elicit human preference comparisons or other feedback on.
This is important as it is expensive to collect the large quantities of high-quality labeled data needed by current approaches.
For example, \textcite{ziegler:2019} collected 60,000 comparisons between summaries of articles, but found the resulting model simply learned to copy the initial sentences. %
They attribute this to labellers under time pressure preferring copies as a fast heuristic for accuracy.
Improved sample efficiency will enable us to collect smaller but higher quality datasets, decreasing the cost of data collection and improving reliability.

Another commonly encountered problem is that RL fine-tuning learns to exploit the learned reward model.
This results in a language model that generates gibberish (but high-reward) outputs~\parencite[Table 10]{ziegler:2019}.
Prior work has avoided this by adding a KL penalty relative to the pre-trained language model as an auxiliary reward.
However, this regularization may limit task performance, especially for tasks that are very different from that seen during pre-training.
Estimating reward model uncertainty allow us to instead perform risk-averse~\parencite{urpi:2021,singh:2020} or risk-constrained~\parencite{chow:2018} reinforcement learning.

In this paper, we investigate whether an ensemble of fine-tuned language models can provide accurate estimates of uncertainty.
Specifically, we randomly reinitialize the final layer of each ensemble member, and then fine-tune the entire network based on human preference comparisons.
Although ensembles are not always the best performing method, they have worked reasonably well in a wide range of applications~\parencite{christiano:2017,lakshminarayanan:2017,beluch:2018,vyas:2018,ovadia:2019,gustafsson:2020}.
Despite this prior track record, we find that ensembles perform poorly in our settings.
In particular, we find \emph{no benefit} from active learning using ensembles compared to randomly sampling data.
Moreover, on a synthetic dataset the ensemble is only slightly more accurate on points with higher confidence (lower variance between ensemble members).

We speculate this negative result is due to limited diversity between ensemble members when fine-tuning from a single pre-trained model.
By contrast, most prior work on uncertainty estimation has focused on training a model from scratch, where this issue does not arise.
But in our setting, it is both more computationally and sample efficient to fine-tune a pre-trained language model, leveraging the natural language representations it has already learned~\parencite{radford:2019,brown:2020,rae:2021}.
Unfortunately, pre-trained language models' notion of uncertainty is for next-token prediction.
Moreover, language models' uncertainty is unreliable out-of-domain~\parencite{desai:2020,kamath:2020}, and is poorly calibrated even in-domain for some tasks such as Q\&A~\parencite{jiang:2021}.

Both active learning and fine-tuning a pre-trained model share the goal of improving sample efficiency.
However, we conjecture there is a crucial tradeoff between them: we must stay close to the pre-trained model to benefit from fine-tuning, but this lack of diversity will hurt uncertainty estimation.
This casts doubt on the dominant paradigm of developing a single large pre-trained model.
We expect that, for the purpose of uncertainty estimation, it may be better to learn multiple distinct smaller pre-trained models, or otherwise introduce uncertainty into the pre-trained model (e.g. via dropout~\parencite{gal:2016}).

\section{Related Work}

Our work seeks to develop techniques for uncertainty estimation and active learning for fine-tuning pre-trained language models.
Uncertainty estimation~\parencite{gawlikowski:2022} and active learning~\parencite{settles:2009} are both well-established fields, while fine-tuning language models is a new but popular area of research~\parencite{bommasani:2021}.
However, to the best of our knowledge the only prior work for uncertainty in fine-tuned language models is limited to calibrating the \emph{total} uncertainty of these models~\parencite{desai:2020,jagannatha:2020,kamath:2020,jiang:2021}.

While calibrated total uncertainty is helpful, many applications (such as active learning) depend on separately estimating aleatoric (or data) uncertainty and epistemic (or model) uncertainty~\parencite{hullermeier:2021}.
Aleatoric uncertainty is due to irreducible randomness in the data generating process.
By contrast, epistemic uncertainty is due to a lack of knowledge and could be reduced by seeing more data points.

Uncertainty estimation for deep learning~\parencite{gawlikowski:2022} seeks to provide well-calibrated estimates of epistemic and aleatoric uncertainty for neural networks.
Ensembles are a simple and popular approach~\parencite{lakshminarayanan:2017}: training multiple models, with their disagreement estimating model uncertainty.
They have been found to be a strong baseline in a variety of applications~\parencite{beluch:2018,vyas:2018,ovadia:2019,gustafsson:2020}, although they have been criticized on both a theoretical~\parencite{osband:2018} and empirical~\parencite{osband:2022} basis.
Hypermodels are a related approach~\parencite{ha:2016,dwaracherla:2020}, generating the weights of another network as a function of some noise variable, effectively simulating a very large ensemble.
Bayesian neural networks are an alternative class of approaches that learn a distribution over the weights of a neural network~\parencite{blundell:2015,gal:2016}.

Active learning~\parencite{settles:2009} seeks to achieve greater sample efficiency by having the algorithm select which data points to ask a human annotator to label.
One effective heuristic selects the point with maximal \emph{epistemic uncertainty}, i.e.\ the point the model is most unsure about, making active learning and uncertainty estimation intricately connected.
This is related to the notion of query-by-committee~\parencite{seung:1992}: selecting points where a ``committee'' of models most disagree, whether a literal ensemble of models or a generative model class such as a hypermodel or Bayesian neural network.

It is, however, possible to perform active learning without any explicit notion of epistemic uncertainty.
For example, uncertainty sampling~\parencite{lewis:1994} simply queries the point which a single model is least confident in the classification of, regardless of whether this is due to epistemic or aleatoric uncertainty.
Moreover, there are also methods that select the points based on the expected change to the model, e.g.~\textcite{settles:2008b}), and those that seek to minimize the variance over model parameters, e.g.~\textcite{zhang:2000}).

In this paper, we focus on learning reward functions from preference comparisons.
This typically requires collecting \emph{de novo} human preference comparisons for the task at hand -- an expensive and time-consuming process.
Prior work has therefore applied active learning to this setting to reduce the size of the dataset required.
\textcite{christiano:2017} propose an ensemble-based method similar to our approach, finding it improves sample efficiency in around three quarters of the continuous control and Atari tasks tested. 
More sophisticated approaches have also been developed, but are difficult to scale and have only been demonstrated in low-dimensional environments~\parencite{sadigh:2017,akrour:2012}.

An alternative way to boost sample efficiency is to use unsupervised learning to leverage vast unlabeled datasets~\parencite{radford:2019,brown:2020}.
These \emph{foundation models} can then be fine-tuned on labeled data to perform a particular task~\parencite{bommasani:2021}.
Crucially, the number of labels required for fine-tuning is far smaller than that of training a model from scratch.

This fine-tuning approach has been successfully applied to learning natural language tasks from preference comparisons.
\textcite{ziegler:2019} fine-tune language models to continue a prompt in a particular style, producing completions that users prefer to the original model 88\% of the time after only 5,000 comparisons.
They also train the model to summarize documents, but find it only learns naive copying behavior.
\textcite{stiennon:2020} refine this work with improved data collection, producing summaries that outperform human reference summaries after less than 100,000 comparisons.

This is a relatively small amount of data, and models can do tolerably well with an order of magnitude less data than this.
The use of a foundation model therefore clearly massively improves sample efficiency.
On the other hand, 100,000 comparisons is still substantial, representing over 4 years of full-time work!\footnote{Assuming 5 minutes per comparison, and 40 hours a week for 52 weeks a year.}
It is therefore tempting to combine active learning with the use of a foundation model, to improve sample efficiency beyond what either method can do alone.
While prior work has attempted to leverage language models to automate part of the labeling process~\parencite{wang:2021,liu:2022}, we are not aware of any work fine-tuning language models that has tried actively selecting points for human feedback.
We seek to fill this gap in the literature in the remainder of the paper.

\section{Uncertainty Estimation}

\subsection{Problem Formulation}
\label{sec:uncertainty-estimation:problem}
We seek to learn a distribution over reward models $R$ given some human feedback dataset $\mathcal{D}$, and pre-trained language model $M$.
We assume the dataset $\mathcal{D}$ consists of preference comparisons between possible document completions, following \textcite{ziegler:2019} and \textcite{stiennon:2020}.
The base model $M(x_k \mid x_1, \cdots, x_{k-1})$ is trained on a large unsupervised dataset to predict the next token $x_k$ completing a sequence $x_1,\cdots,x_{k-1}$.

A key problem we face is that the human feedback dataset $\mathcal{D}$ is typically rather small.
Certainly the feedback dataset $\mathcal{D}$ is far smaller than available unsupervised training corpora. %
Practical learning procedures must therefore make extensive use of the pre-trained model $M$, such as by using it as an initialization for $R$ and then fine-tuning the parameters.

\subsection{Ensemble Models}
\label{sec:uncertainty-estimation:ensemble}

To approximate a distribution, we use an ensemble of $n$ reward models $\{R_1,\cdots,R_n\}$.
We construct each model $R_i$ by following a similar approach to \textcite{ziegler:2019} and \textcite{stiennon:2020}.
Specifically, we copy the weights of the pre-trained model $M$ and
replace the final layer with a randomly initialized linear layer that outputs a reward (a real number).
The entire network $R_i$ is then fine-tuned based on human preference comparisons, where the reward output is used as logits in a logistic regression model (i.e.\ higher reward documents are exponentially more likely to be chosen).

Diversity between the ensemble members comes from two sources: initialization of the final layer, and bootstrap sampling of the dataset.
The final layer of each ensemble member $R_i$ is randomly initialized with a different random seed.
Moreover, since we fine-tune the \emph{entire} network, parameters at \emph{all} layers will tend to differ between ensemble members at convergence.

Additionally, the dataset of each ensemble member is constructed via bootstrap sampling.
Specifically, for each data point and ensemble member, we independently sample a weight of $2$ or $0$ with 50\% probability in the first epoch and reuse these weights in subsequent epochs.
This is analogous to building a dataset for each ensemble member by independently sampling points (without replacement).

\subsection{Active Learning}
\label{sec:uncertainty-estimation:active}

We use the ensemble model for active learning in two ways.
First, we use Thompson sampling~\parencite{thompson:1933} that picks an ensemble member uniformly at random, and then selects the data point which that ensemble member predicts the highest reward for.
Second, we select points with maximal variance between the predictions of each ensemble member, an estimate of \emph{epistemic uncertainty}.
Specifically, since the comparisons are \emph{binary}, we compute the variance over the ensemble members' probabilities that the first summary will be selected over the second.

\section{Experiments}

We evaluate the quality of our uncertainty estimates by using active learning on real human data.
Active learning is a key application of reward model uncertainty, and is also widely used to evaluate uncertainty estimates~\parencite[Table~IV]{gawlikowski:2022}.
In contrast to prior work in other domains~\parencite{christiano:2017,beluch:2018}, we find no significant improvement from using active learning with our ensemble model.

We also evaluate on a synthetic dataset where we know the aleatoric uncertainty exactly.
We find the estimated epistemic uncertainty is moderately predictive of model error, and the correlation increases in larger ensembles.
However, the correlation is weak even in the best case scenario, likely explaining the failure of active learning.
This suggests the need for further research to improve uncertainty estimates in this domain.

We describe our experimental setup in the next section. Subsequently, we present the results on active learning, and finally evaluate the estimated epistemic uncertainty quality on the synthetic dataset.

\subsection{Experimental Setup}

We train and evaluate reward models on a dataset of human preference comparisons between automatically generated summaries collected by \textcite{stiennon:2020}.
The documents summarized are from the TL;DR dataset of Reddit posts~\parencite{volske:2017} and the CNN/DM dataset of news articles~\parencite{hermann:2015}.
The TL;DR comparisons are split into a training (92,858 comparisons), validation (33,083 comparisons) and test (50,719 comparisons) set.
The 2,284 CNN/DM comparisons are used as an out-of-distribution test set to check the robustness of the reward model's predictions and uncertainty estimates.

Our pre-trained language model $M$ is a
\if@submission
---details removed for anonymization---
\else
small version of Gopher~\parencite{rae:2021} 
\fi
trained on the C4 dataset~\parencite{raffel:2020}.
The model architecture is similar to GPT-2~\parencite{radford:2019}, but uses root-mean-square layer normalization~\parencite[RMSNorm]{zhang:2019} instead of layer normalization~\parencite{ba:2016} and relative positional encoding~\parencite{shaw:2018} instead of absolute positions (see section~\ref{sec:supp:pre-trained} for details).
Our primary model has 417M parameters, but we also test 117M and 1.3B parameter models in some experiments to evaluate dependence on model size.

\subsection{Active Learning}
\label{sec:experiments:active}

We test our 417M parameter model in a pool-based active learning setup on the TL;DR training dataset.
We give the algorithm access to the full unlabeled dataset, and allow it to choose a total of 4,096 data points to reveal the human feedback label for.
Specifically, we sample 16 preference comparisons at random, and then choose the one estimated to be the most informative.
Training in subsequent epochs replays shuffled data from the first epoch.

This pool-based setup has the benefit of allowing methods to be evaluated purely offline, on realistic data.
However, an important caveat is that it may underestimate the benefit of active learning.
In particular, synthesizing novel pairs of summaries might produce more informative comparison pairs than those naturally present in the dataset.

\begin{figure}
    \centering
    \includegraphics{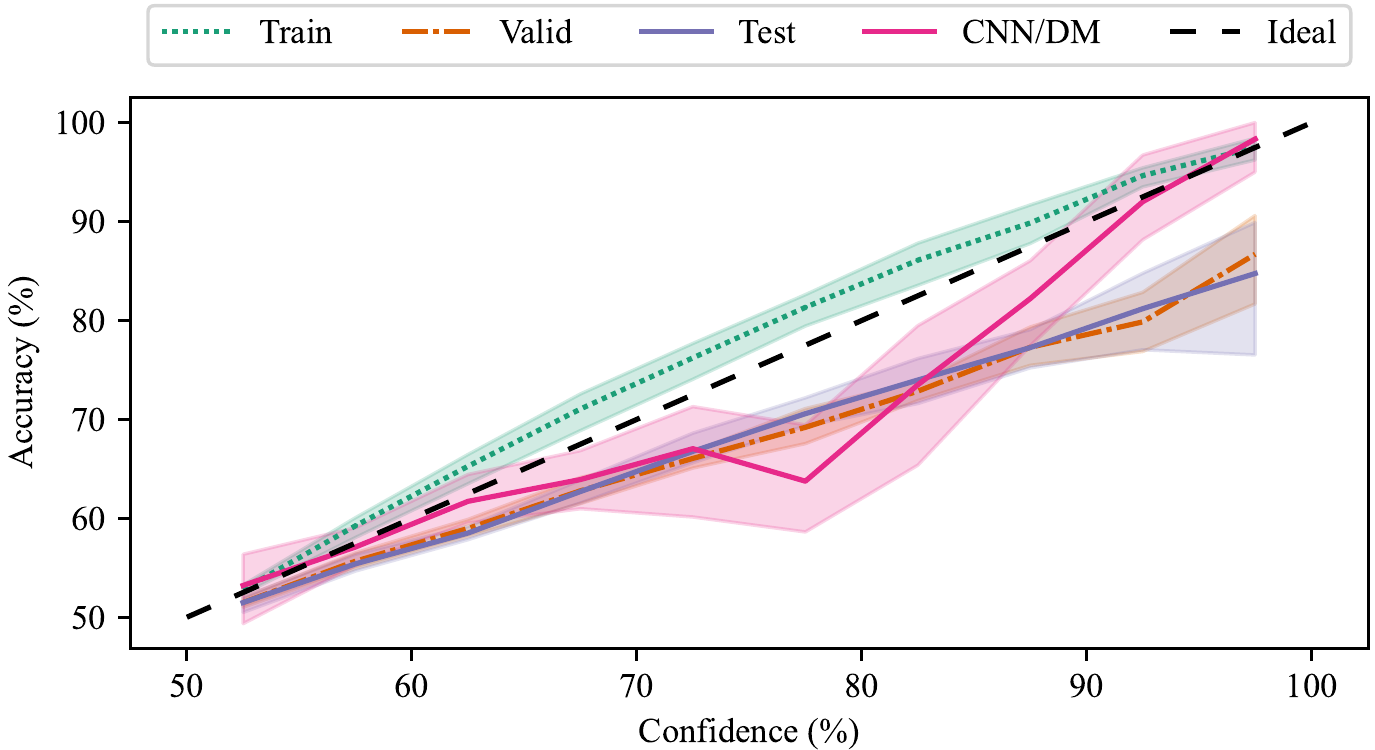}
	\caption{The probability of one summary being preferred over another is well-calibrated for the 417M parameter model trained on the full dataset. Figure~\ref{fig:supp:calibration} shows that calibration improves with model and dataset size. We compute summary selection probability \calibrationlegendcommon{}} 
    \label{fig:eval-softmax-calibration}
\end{figure}

\begin{figure}
    \centering
    \includegraphics{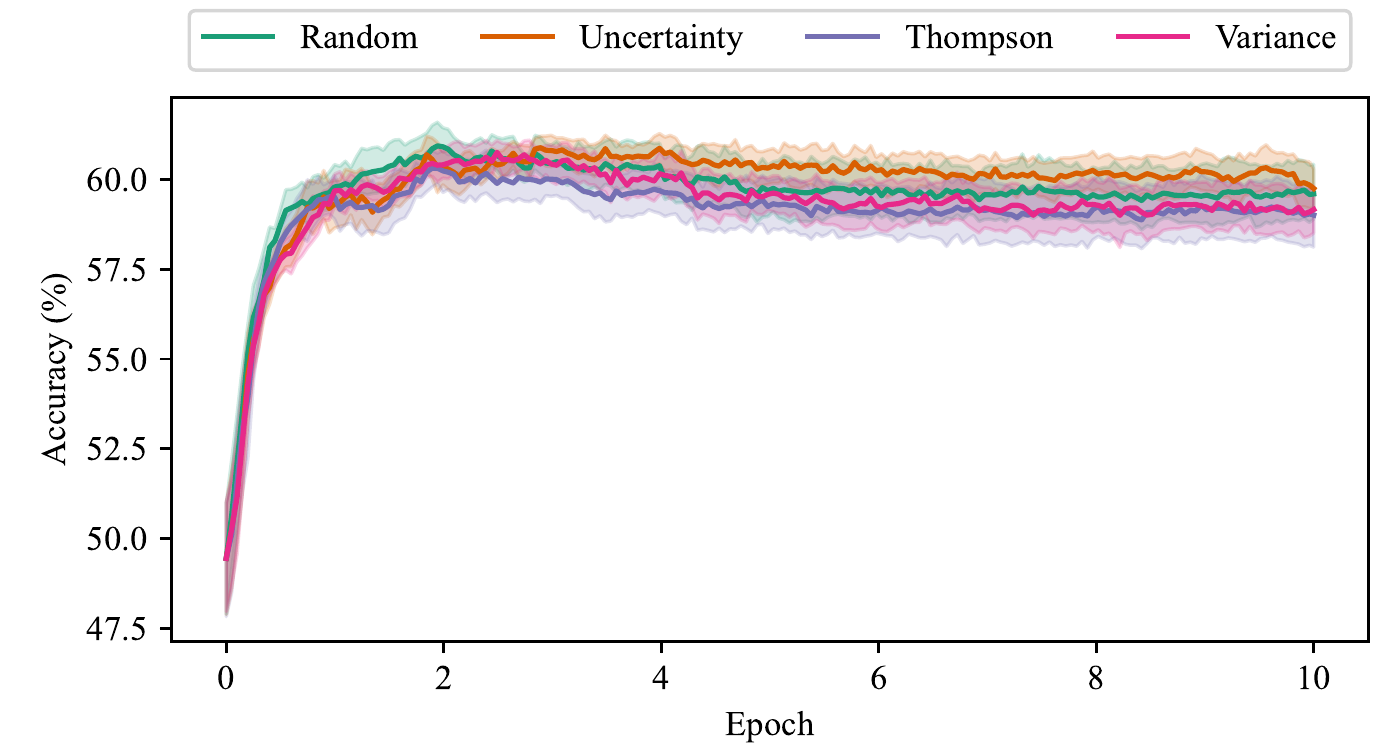}
    \caption{Active learning methods do not significantly outperform \randomkey{}. We report accuracy on the \redditlongform{} dataset while training a 417M parameter model on 4,096 points from the TL;DR Train dataset. \activelegendcommon{}} 
    \label{fig:eval-active-learning}
\end{figure}

We use uncertainty sampling~\parencite{lewis:1994} as a baseline.
That is, we query labels for the points where the model is the least confident about the class, i.e.\ the \emph{class} probability is closest to uniform.
While simple, this has been found to be a strong baseline~\parencite{settles:2008,yang:2018,gissin:2019}.
Moreover, Figure~\ref{fig:eval-softmax-calibration} shows that these class probabilities are reasonably well calibrated.
However, in Figure~\ref{fig:eval-active-learning}, we find that uncertainty sampling (\uncertaintykey{}) does not significantly outperform random (\randomkey{}) selection.

We conjecture that this is because uncertainty sampling favors points with the highest total uncertainty.
However, points with high aleatoric uncertainty will tend to be \emph{less} informative as the human feedback is itself highly stochastic.
In many datasets, aleatoric uncertainty is comparable across points, in which case optimizing for total uncertainty is similar to optimizing for epistemic uncertainty.
However, in our case, aleatoric uncertainty is highly variable between points (heteroscedastic) and often large enough to drown out epistemic uncertainty.

Fortunately, use of an ensemble model lets us disentangle \emph{epistemic uncertainty}, represented by disagreement between ensemble members, from \emph{aleatoric}, estimated by the aggregate prediction of the ensemble.
We leverage this using Thompson sampling (\thompsonkey{}) and by selecting points with maximal variance (\variancekey{}), as described in Section~\ref{sec:uncertainty-estimation:active}.
Perhaps surprisingly, we find in Figure~\ref{fig:eval-active-learning} that these more sophisticated methods also do not outperform \randomkey{} sampling.

Notably \textcite{christiano:2017} reported positive results for active preference learning in simulated robotics and Atari games using a similar approach to \variancekey{}.
However, they did find the performance was highly variable between environments.
This is consistent with prior results showing that active learning performance is heavily dependent on the dataset~\parencite{mussmann:2018}.

\subsection{Oracle Labeler}
\label{sec:eval:oracle}

\begin{figure}
    \begin{subfigure}{\textwidth}
        \centering
	\includegraphics{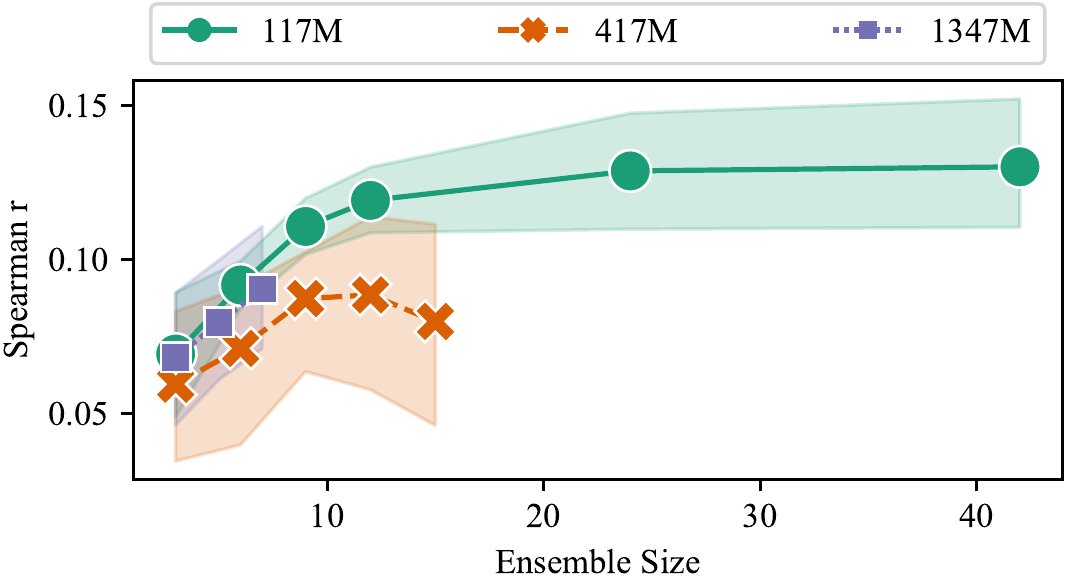}
	\caption{TL;DR Test Dataset, with bootstrapping.}
        \label{fig:eval-oracle-corr:reddit}
    \end{subfigure}
    \begin{subfigure}{\textwidth}
        \centering
	\includegraphics{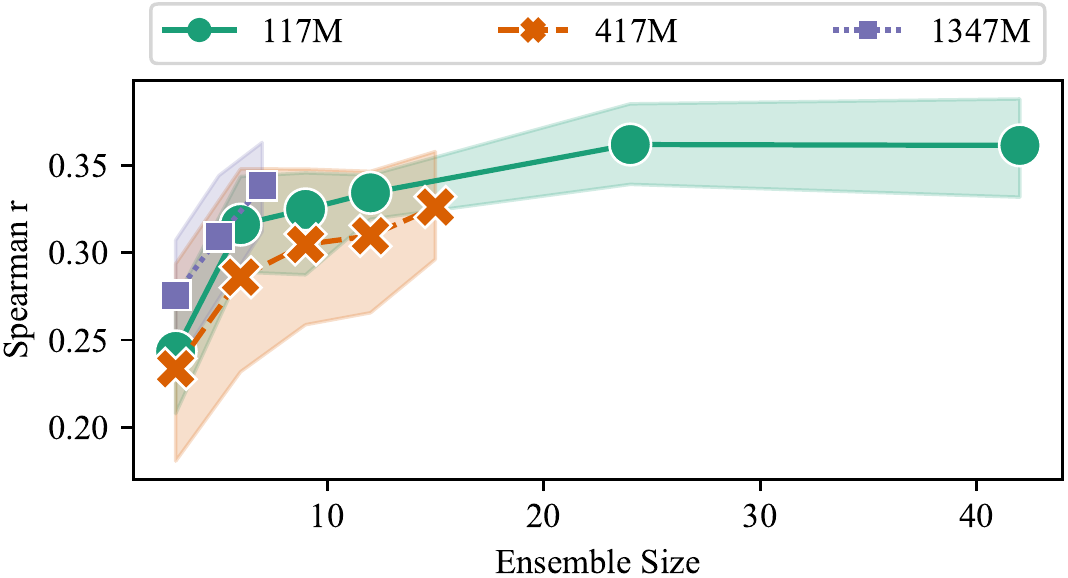}
	\caption{Out-of-distribution CNN/DM Dataset, with bootstraping.}
        \label{fig:eval-oracle-corr:cnndm}
    \end{subfigure}
    \begin{subfigure}{\textwidth}
        \centering
	\includegraphics{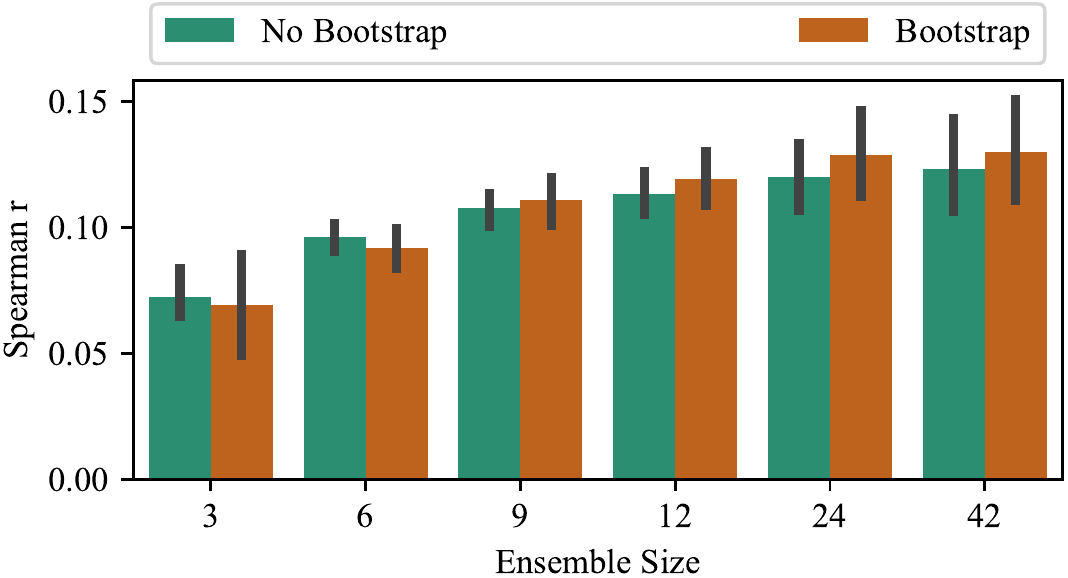}
	\caption{117M parameter model on the TL;DR Test dataset, \bootstrapkey{} and \nobootstrapkey{} bootstrapping.}
        \label{fig:eval-oracle-corr:bootstrap}
    \end{subfigure}
    \caption{We report the Spearman correlation between model error and estimated epistemic uncertainty ($y$-axis), against the number of ensemble members ($x$-axis). The best model has at most $r = 0.36$ correlation, indicating the uncertainty estimates are unreliable. \textbf{(a), (b)}: The correlation increases modestly with the number of ensemble members and model parameter count. \textbf{(c)}: The correlation tends to be higher with bootstrapping, but there is no significant difference as the confidence intervals (error bars) overlap. \uncertaintylegendtail{} See Figures~\ref{fig:supp:eval-oracle-corr-bootstrap:valid1}, \ref{fig:supp:eval-oracle-corr-bootstrap:valid2_reddit} and~\ref{fig:supp:eval-oracle-corr-bootstrap:valid2_cnndm} for results on other models and datasets.}
    \label{fig:eval-oracle-corr}
\end{figure}

In the previous section we found that active learning using a bootstrapped ensemble does not outperform random sampling.
However, this negative result could be caused by several factors.
For one, the estimate of epistemic uncertainty could be low quality, despite the \emph{total} uncertainty being well-calibrated per Figure~\ref{fig:eval-softmax-calibration}.
But alternately, perhaps the acquisition function (how the uncertainty estimates are translated into a score for active learning) could be at fault.
A further possibility is that different points in the dataset are similarly informative -- in that case, no active learning method could substantially outperform random sampling.

To directly evaluate our uncertainty estimates we test against an oracle labeler that was trained on the full human feedback dataset.
We then train an ensemble model on a subset of this data, with labels sampled from predictions of the oracle.
The oracle therefore gives us the ground-truth distribution over labels: in other words, the aleatoric uncertainty of the data is known exactly.
We can use this to more precisely evaluate the quality of our epistemic uncertainty.

In particular, we estimate epistemic uncertainty based on the variance in predictions between ensemble members, in the same way as for the ``max variance'' active learning approach in the previous section.
For each data point, we compute the model error as measured by the KL divergence between the model's predicted distribution and the oracle's predicted distribution.
We then compute the Spearman (rank) correlation between the model error and estimated epistemic uncertainty.
The higher the correlation, the more the model ``knows what it knows'': while it still often make mistakes, it is aware of the points where it is likely to do so.
By contrast, with low correlation, the model is sometimes highly confident about a point that it makes an error on.

In Figures~\ref{fig:eval-oracle-corr:reddit} and~\ref{fig:eval-oracle-corr:cnndm}, we report this Spearman correlation measuring epistemic uncertainty quality ($y$-axis) for ensembles with varying numbers of members ($x$-axis) and parameter count (line color).
We find that the correlation is fairly low in all cases, reaching at most $r=0.36$ in Figure~\ref{fig:eval-oracle-corr:cnndm}, explaining only $r^2 \approx 0.13$ of the variance.
However, the correlation does increase with ensemble size, from $r=0.25$ at the lower end with an ensemble size of $3$ to $r=0.36$ with an ensemble size of $42$.

Bootstrapping slightly improves mean uncertainty quality, but the difference is not statistically significant, and the size of the ensemble is more important.
In Figure~\ref{fig:eval-oracle-corr:bootstrap} we observe the uncertainty quality with bootstrapping (orange bars) are generally above the green bars (without bootstrapping) for the 117M parameter model, but that the 95\% confidence intervals (error bars) are overlapping.
We find similar results for the 417M and 1347M parameter models in Figure~\ref{fig:supp:eval-oracle-corr-bootstrap:valid2_reddit} in the appendix.

Overall, these results indicate that the poor active learning performance of the bootstrapped ensemble in Section~\ref{sec:experiments:active} is largely due to low quality uncertainty estimates.
While it is possible that better performance could be obtained with an impoved acquisition function, the success of \variancekey{} in prior work~\parencite{christiano:2017} makes it unlikely that this choice alone explains the negative result.
However, our results may be sensitive to the choice of dataset, which can affect \emph{both} the uncertainty quality and the strength of random sampling.
A promising direction for future work is to evaluate this approach in a broader range of natural language tasks in case the performance in summarization is unusually poor.

\section{Discussion}

We investigated the use of ensembles and bagging to estimate uncertainty for pre-trained language models fine-tuned for summarization.
Despite ensembles working well in a range of applications~\parencite{lakshminarayanan:2017,beluch:2018,vyas:2018,ovadia:2019,gustafsson:2020}, we find no benefit from active learning using ensembles in our setting.
Moreover, evaluating against an oracle labeler we find that the uncertainty estimation -- while better than random -- is of low quality, explaining at most 13\% of the variance in model error.

There are two key limitations of this work.
First, we have only investigated uncertainty estimation in a single task: learning to discriminate between good and bad summaries of documents.
While the dataset does span a range of topics (e.g.\ business, sport, politics), the task is still relatively narrow compared to, for example, open-ended dialogue.
Consequently, it may be that random sampling is a fairly strong baseline in this task.
An interesting direction for future work would be to estimate an upper bound for active learning performance in this task, for example by greedily selecting data points that most improve accuracy on held-out data.

Second, there are a variety of other methods for uncertainty estimation that we have not yet attempted.
One approach we find particularly promising are linear hypermodels~\parencite{dwaracherla:2020}: learning a linear mapping from a noise source to neural network parameters, that are then used for inference.
Although our modestly sized ensembles did not achieve adequate uncertainty quality, extrapolating from Figure~\ref{fig:eval-oracle-corr} it seems that ensembles of $1000$s of models might be sufficient.
Hypermodels provide a space-efficient way of approximating this.
An alternative approach would be to fine-tune only the biases, similar to BitFit~\parencite{zaken:2021}, enabling many more ensemble members to fit into limited memory.

In recent years, machine learning has moved towards large, pre-trained \emph{foundation models}~\parencite{bommasani:2021}.
Fine-tuning these models is often significantly more sample efficient than training from scratch.
However, our work highlights this benefit does not come for free: staying close to the foundation model ``prior'' will cause a lack of diversity in ensemble members (or hypermodel samples), yielding poor quality uncertainty estimates.
We believe it will either be necessary to modify foundation model training procedures to include uncertainty estimation, such as by training ensembles of foundation models, or develop new methods for uncertainty estimation that can tolerate this lack of diversity.

\clearpage

\subsection*{Acknowledgements}
\if@submission
Removed for anonymization.
\else
Thanks to Vladimir Mikulik for running several of the experiments; to John Aslanides, Francis Song, Roman Ring and Nat McAleese for code-review and other technical assistance; and to Sebastian Farquhar for various helpful discussions.
\fi

\printbibliography

\clearpage

\appendix

\section{Supplementary Material}

\subsection{Comparison to Prior Work}
We find that our language and reward models successfully replicate prior work~\parencite{radford:2019,stiennon:2020}. Although there are some minor implementation differences, discussed below, these do not seem to significantly change the overall language modelling perplexity or reward modelling accuracy.
Accordingly, we expect our results to be applicable to other pre-trained language models and preference comparison algorithm implementations.

\subsection{Pre-trained Language Models}
\label{sec:supp:pre-trained}
\if@submission
Our experiments use models trained on the ``Colossal Clean Crawled Corpus''~(C4)~\parencite{raffel:2020}, derived from the Common Crawl\footnote{\url{https://commoncrawl.org}}. 
Our model architecture is similar to GPT-2~\parencite{radford:2019}, with the following differences:
\else
Our experiments use small models from the Gopher family~\parencite{rae:2021} trained on the ``Colossal Clean Crawled Corpus''~(C4)~\parencite{raffel:2020}, derived from the Common Crawl\footnote{\url{https://commoncrawl.org}}. 
In common with Gopher, our model architecture is similar to GPT-2~\parencite{radford:2019}, with the following differences:
\fi
\begin{itemize}
    \item The models use root-mean-square layer normalization~\parencite[RMSNorm]{zhang:2019} instead of layer normalization~\parencite[LayerNorm]{ba:2016}.
    \item The models use relative positional encodings~\parencite{shaw:2018} instead of absolute positions.
\end{itemize}

Our pre-trained models achieve a similar performance level as comparably sized GPT-2 models, summarized in Table~\ref{tab:supp:pre-trained-eval}.
In particular, our models achieve lower (better) perplexity on WikiText 103 than GPT-2 models.
However, our models have an 8-13\% lower (worse) LAMBADA accuracy than reported for GPT-2.
We believe this difference is largely due to GPT-2 models using a stop word filter, which increased accuracy by around 11\%~\parencite[section 3.3]{radford:2019}.
Another confounder is that GPT-2 was evaluated on a raw (unprocessed) version of LAMBADA~\parencite{yaroslavb:2019}, whereas we evaluate on the standard dataset.

\begin{table}
    \centering
    \begin{tabular}{llll}
        \toprule
        \textbf{Model} & \textbf{Parameter} & \textbf{LAMBADA} & \textbf{WikiText 103} \\
        \textbf{Source} & \textbf{Count} & \textbf{Accuracy (\%)} & \textbf{Perplexity} \\
        \midrule
        Ours & 117M & 34.3 & 33.03 \\
        Ours & 417M & 42.7 & 23.90 \\
        Ours & 1374M & 54.1 & 17.053 \\
        \midrule
        GPT-2 & 117M & 45.99 & 37.50 \\
        GPT-2 & 345M & 55.48 & 26.37 \\
        GPT-2 & 1542M & 63.24 & 18.34 \\
        \bottomrule
    \end{tabular}
	\caption{Our pre-trained language models achieve similar performance to comparably sized GPT-2 models on WikiText 103~\parencite{merity:2016}. They perform around 8-13\% worse on LAMBADA accuracy than reported for GPT-2, however we believe this is due to GPT-2 using a stop word filter and an unprocessed version of LAMBADA; see section~\ref{sec:supp:pre-trained}.}
    \label{tab:supp:pre-trained-eval}
\end{table}

\subsubsection{Reward Models}

\begin{table}[t]
    \centering
    \begin{tabular}{llll}
        \toprule
        \textbf{Model} & \textbf{Parameter} & \textbf{Evaluation} & \textbf{Labeler} \\
        \textbf{Source} & \textbf{Count} & \textbf{Dataset} & \textbf{Agreement (\%)} \\
        \midrule
        Ours & 117M & \redditlongform{} & $60.71\% \pm 2.11\%$ \\
        Ours & 117M & \cnndmlongform{} & $58.68\% \pm 3.56\%$ \\
        Ours & 417M & \redditlongform{} & $61.84\% \pm 3.00\%$ \\
        Ours & 417M & \cnndmlongform{} & $61.37\% \pm 4.99\%$ \\
        Ours & 1347M & \redditlongform{} & $64.84\% \pm 1.63\%$ \\
        Ours & 1347M & \cnndmlongform{} & $65.90\% \pm 3.27\%$ \\
        \midrule
        \textcite{stiennon:2020} & 1.3B & 1.3B supervised & $65.8\% \pm 2.0\%$ \\
        \textcite{stiennon:2020} & 1.3B & 6.7B supervised & $67.5\% \pm 1.1\%$ \\
        \textcite{stiennon:2020} & 1.3B & 6.7B feedback & $57.4\% \pm 2.0\%$ \\
        \textcite{stiennon:2020} & 6.7B & 1.3B supervised & $70.8\% \pm 1.8\%$ \\
        \textcite{stiennon:2020} & 6.7B & 6.7B supervised & $69.7\% \pm 1.1\%$ \\
        \textcite{stiennon:2020} & 6.7B & 6.7B feedback & $62.3\% \pm 2.1\%$ \\
        \bottomrule
    \end{tabular}
	\caption{Our reward models fine-tuned on the human feedback dataset from \textcite{stiennon:2020} achieve similar accuracy (agreement with human labels) as their models. Unfortunately we cannot perform a direct comparison as \textcite{stiennon:2020} only report on labeler agreement for subsets of the data that we are unable to reproduce. However, we find that our 1347M model achieves an accuracy of $64.84\% \pm 1.63\%$ that is well above the worst figure, $57.4\% \pm 2.0\%$, reported by \textcite[Table~20]{stiennon:2020} for their 1.3B model, but somewhat below the best figure, $67.5\% \pm 1.1\%$. For each model, we evaluate the checkpoint with the highest validation accuracy. We report $95\%$ confidence intervals computed using a $t$-distribution based on the mean and standard deviation across 5 seeds.}
    \label{tab:supp:reward-model-full}
\end{table}

In Table~\ref{tab:supp:reward-model-full} we report the accuracy of reward models trained on the full human feedback dataset from \textcite{stiennon:2020}, observing comparable accuracy to their models.
Although the focus of our work is on introducing uncertainty estimation and active learning, and not improving on accuracy in existing settings, it is pleasing that we were able to easily replicate prior work.
This both validates our training pipeline, and provides additional confirmation of the results in \textcite{stiennon:2020}.
Unfortunately we were unable to make a head-to-head comparison due to a lack of information as to the subset of the dataset that \textcite{stiennon:2020} report on.
However, our accuracy seems comparable to theirs across a range of dataset slices.

\subsection{Model Calibration}

\begin{figure}
    \centering
    \includegraphics{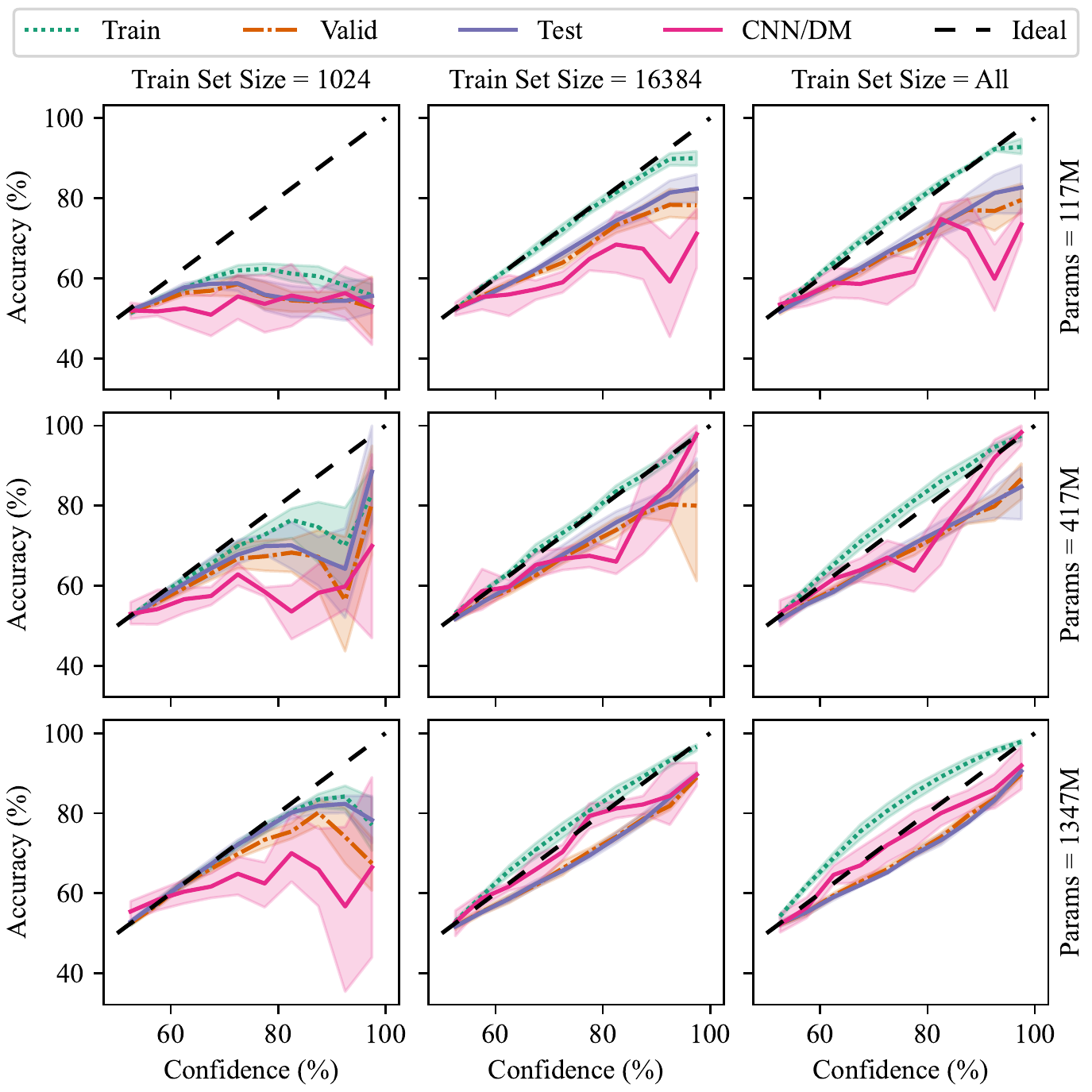}
    \caption{The models' probabilities of one summary being preferred over another are well-calibrated, and calibration improves with dataset and model size. We compute summary selection probability for each model (rows) and training dataset size (columns) \calibrationlegendcommon{}}
    \label{fig:supp:calibration}
\end{figure}

In Figure~\ref{fig:supp:calibration}, we test the degree to which our models are \emph{calibrated} by plotting how often the model was correct ($y$-axis) against its predicted probability ($x$-axis). We find that overall the models tend to be well-calibrated, with lines close to the ideal $y=x$. Calibration is better in models with larger parameter counts, and those that were trained with more data. 

\subsection{Active Learning}

\begin{figure}
    \centering
    \includegraphics{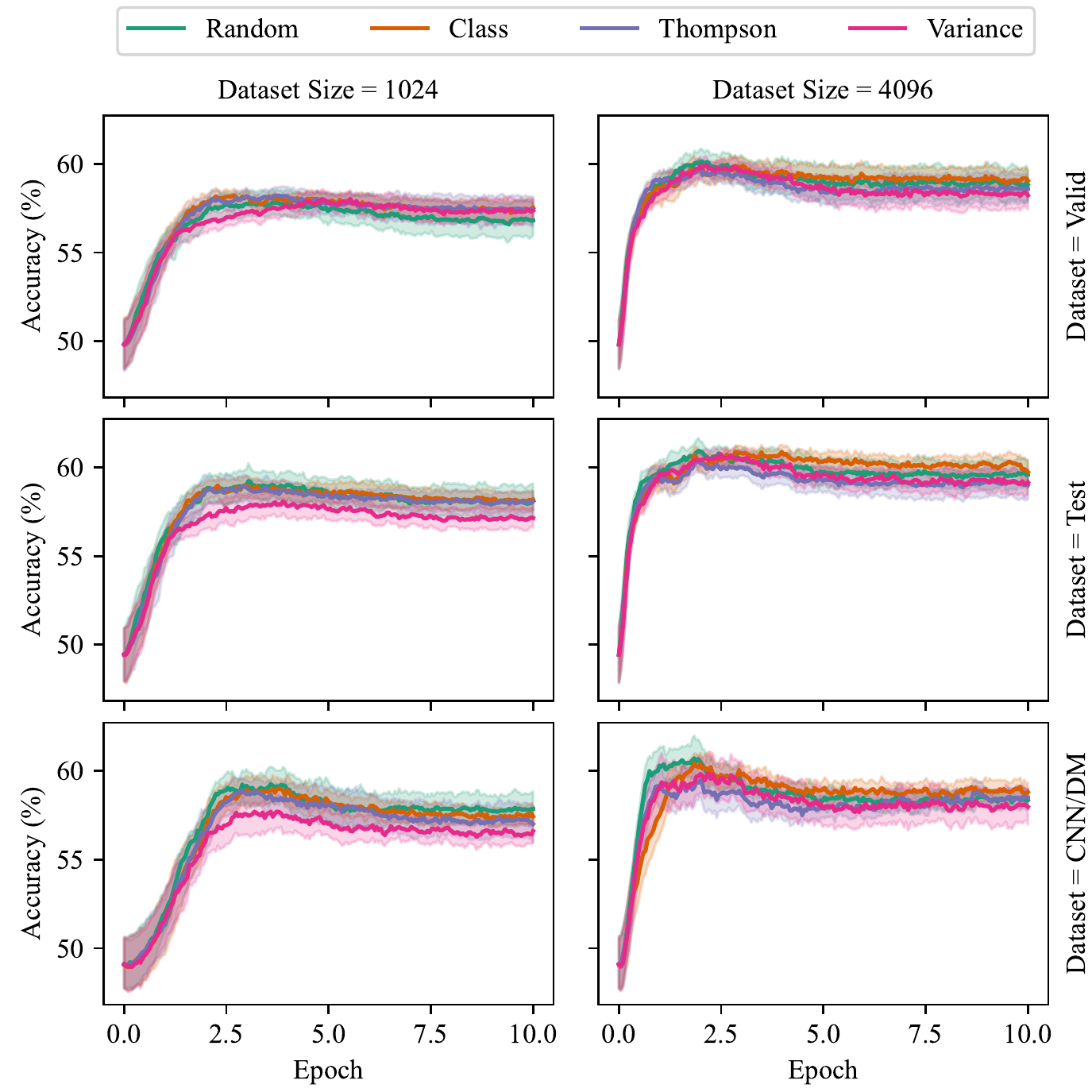}
	\caption{We report accuracy for models trained with active learning on 1,024 (left) or 4,096 (right) data points. The models are trained on a single dataset, but evaluated on three different datasets: TL;DR Valid (top), Test (middle) and out-of-distribution CNN/DM (bottom). No method significantly outperforms \randomkey{} sampling. \activelegendcommon{}}
    \label{fig:supp:eval-active-learning}
\end{figure}

In Figure~\ref{fig:supp:eval-active-learning}, we evaluate the accuracy of different active learning methods (colored lines) trained with different dataset sizes (columns) and evaluated on different test datasets (rows).
In the first epoch, each data point is actively selected; in subsequent epochs, the first epoch is shuffled and replayed.
We find that no method is significantly better than random selection in this setting.

\subsection{Oracle Labeler}

To directly evaluate our uncertainty estimates, we tested their predictions against an \emph{oracle} labeler as described in Section~\ref{sec:eval:oracle}.
This lets us determine if the poor active learning performance is attributable to issues with the uncertainty estimates, or is caused by another factor.
In Figures~\ref{fig:supp:eval-oracle-corr-bootstrap:valid1}, \ref{fig:supp:eval-oracle-corr-bootstrap:valid2_reddit} and~\ref{fig:supp:eval-oracle-corr-bootstrap:valid2_cnndm}, we report the Spearman r correlation coefficient between model error (KL divergence between model and oracle predictions) and estimated epistemic uncertainty (variance between ensemble members) across the TL;DR Validation and Test datasets, and the out-of-distribution CNN/DM dataset.
A higher correlation indicates the model is better calibrated about what it knows: while it still makes mistakes, it can predict the points where it is likely to do so.

\newcommand{\syntheticraterfig}[2]{%
\begin{figure}
    \centering
    \includegraphics{figs/synthetic_rater/bootstrap_grid_#1}
    \caption{\uncertaintylegendhead{} on the #2 dataset. \uncertaintylegendtail{}}
    \label{fig:supp:eval-oracle-corr-bootstrap:#1}
\end{figure}
}

\syntheticraterfig{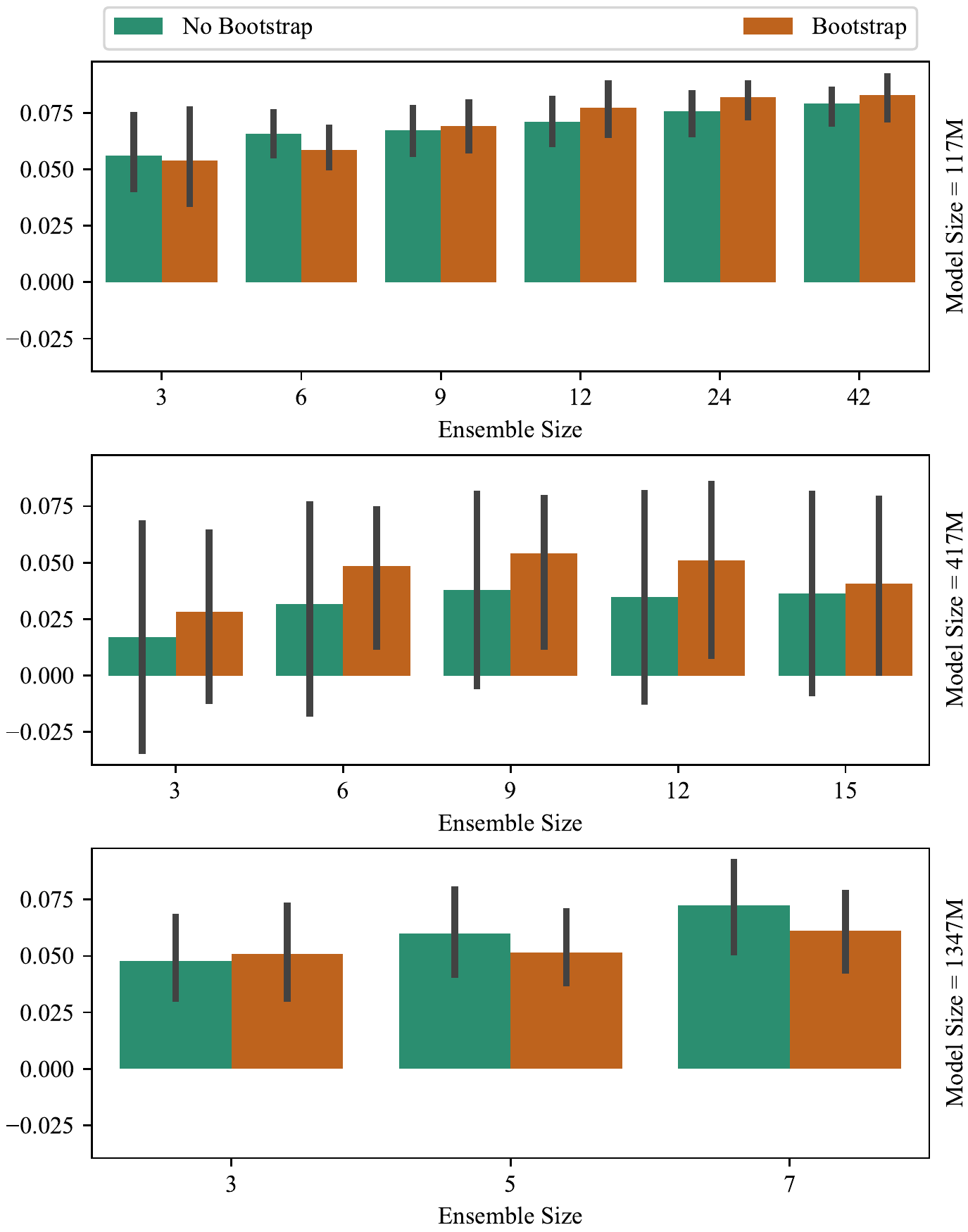}{\validlongform{}}
\syntheticraterfig{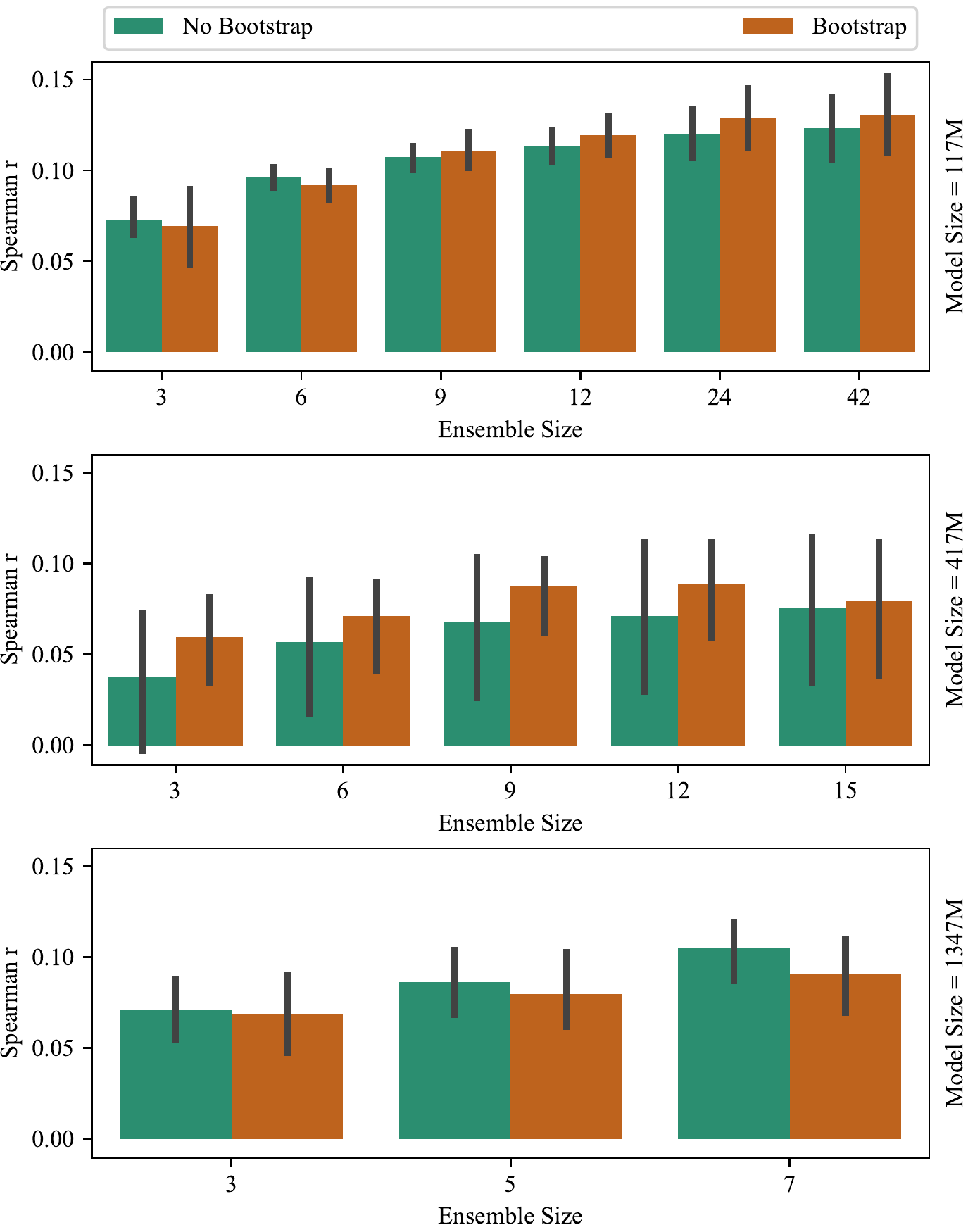}{\redditlongform{}}
\syntheticraterfig{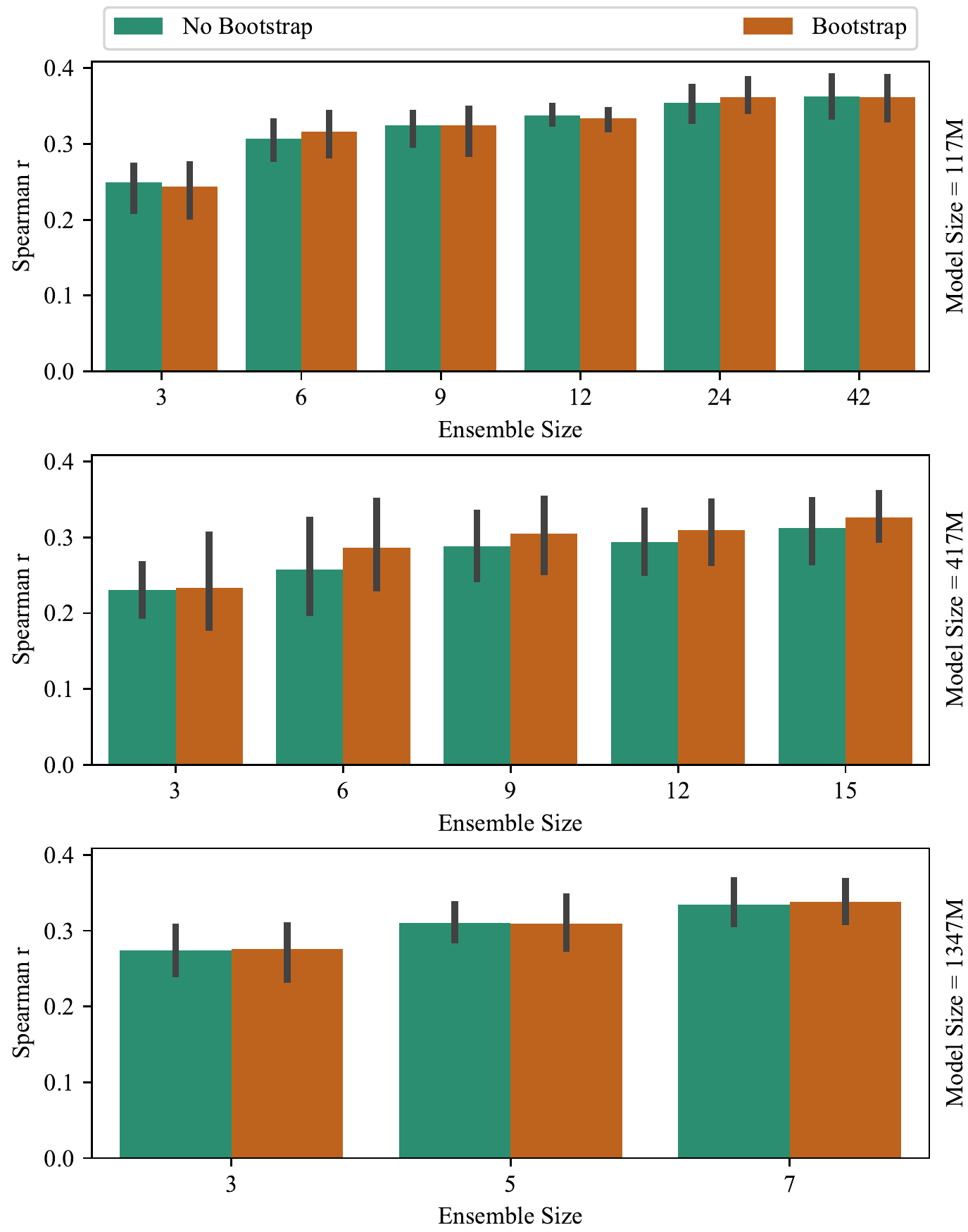}{\cnndmlongform{}}

\end{document}